\documentclass{article}

\usepackage{amsmath,graphicx,mlspconf}

\usepackage{cite}
\usepackage{amsmath,amssymb,amsfonts}
\usepackage{algorithmic}
\usepackage{graphicx}
\usepackage{textcomp}
\usepackage{xcolor}

\newtheorem{theorem}{\bf Theorem}
\newtheorem{lemma}{\bf Lemma}

\newtheorem{corollary}{\bf Corollary}
\newtheorem{proposition}{\bf Proposition}

\title{Generic Bounds on the Maximum Deviations in Sequential Prediction:\\An Information-Theoretic Analysis}
\name{Song Fang, Quanyan Zhu\thanks{This work was supported in part by NSF under grant ECCS-1847056 and SES-1541164, in part by a U. S. DOT grant through C2SMART Center at NYU, and in part by the U.S. DHS through the CIRI under Grant 2015-ST-061-CIRC01.}}
\address{Department of Electrical and Computer Engineering,	New York University, USA\\
	song.fang@nyu.edu, quanyan.zhu@nyu.edu}

\begin{document}
%

\maketitle
\begin{abstract}
In this paper, we derive generic bounds on the maximum deviations in prediction errors for sequential prediction via an information-theoretic approach. The fundamental bounds are shown to depend only on the conditional entropy of the data point to be predicted given the previous data points. In the asymptotic case, the bounds are achieved if and only if the prediction error is white and uniformly distributed.
\end{abstract}
\section{Introduction}
\label{sec:intro}

Nowadays machine learning techniques are becoming more and more prevalent in real-time systems such as real-time signal processing, feedback control, and robotics systems. In such systems, on one hand, decisions on the actions are to be made in a sequential manner (sequential decision making); on the other hand, dynamics of the systems as well as the environment that are determined by physical laws will play an indispensable role and must be taken into consideration (interaction with real world). In this trend, it is becoming more and more critical to be fully aware of the performance limits of the sequential machine learning algorithms that are to be embedded in the autonomous systems operating in real world, especially in scenarios where worst-case performance guarantees are required and must be strictly imposed. 

Sequential prediction (or sequence prediction) \cite{sun2001sequence, dietterich2002machine, cesa2006prediction, shalev2012online, rakhlin2014statistical} has been an important component of sequential learning. In this paper, we will utilize information theory to analyze the fundamental performance bounds of sequential prediction. Information theory \cite{Cov:06} was originally developed to analyze the fundamental performance limitations of communication systems, characterizing any systems that might involve information transmission from one point to another. In a broad sense, the machine learning approaches may be viewed as information transmission processes, as if extracting as much ``information" as possible out of the training data points (cf. discussions in, e.g., \cite{tishby2015deep, shwartz2017opening}) and then transmitting the information to the test data points, so as to reduce as much as possible the ``uncertainty" contained in the latter. In sequential machine learning, this ``information extraction $\to$ information transmission $\to$ uncertainty reduction" process is done in a sequential manner. By virtue of this obvious similarity on the general mechanism level, in this paper we aim to examine the fundamental performance limits of sequential prediction via an information-theoretic approach.


%
%


In linear prediction theory \cite{kailath1974view, makhoul1975linear, pourahmadi2001foundations, vaidyanathan2007theory, caines2018linear}, which has been an important branch of signal processing, the Kolmogorov--Szeg\"o formula \cite{Pap:02, vaidyanathan2007theory, FangACC18, FangCDC17} and Wiener--Masani formula \cite{lindquist2015linear, chen2018role, FangACC18} provide fundamental bounds on the variances of prediction errors for the linear prediction of Gaussian sequences. In this paper, however, we go beyond the linear Gaussian case, since learning algorithms are certainly not restricted to being linear and the noises are oftentimes non-Gaussian. More importantly, we consider the worst-case scenario by minimizing the maximum deviations rather than the variances of the prediction errors \cite{FangITW19arxiv}. In fact, in cases where the variances of the prediction error are minimized, it is possible that the probability of having an arbitrary large deviation in the prediction error is non-zero; for instance, when the prediction error is Gaussian \cite{linearestimation}. This could cause severe consequences in safety-critical systems interacting with real world.

More specifically, we derive generic bounds on the maximum deviations in prediction errors for sequential prediction, by investigating the underlying entropic relationships of the data points composing the sequences. 
The fundamental bounds are applicable to any learning algorithms, while the sequences to be predicted can be with arbitrary distributions. The bounds are characterized explicitly by the conditional entropy of the data point to be predicted given the previous data points. 
Moreover, it is shown that the necessary and sufficient condition for achieving the lower bounds asymptotically is that the prediction error is asymptotically white uniform. This mandates that the optimal (in the sense of minimizing the maximum deviation in the prediction error) learning algorithm is prediction error uniformizing-whitening. In a broad sense, the prediction error being white uniform means there is no useful ``information" remaining in the residue; in other words, all the information that may be utilized to reduce the deviation in the prediction error has been extracted in the uniformizing-whitening procedure.



The remainder of the paper is organized as follows. Section~2 introduces the technical preliminaries. In Section~3, we introduce the information-theoretic bounds for estimation. Section~4 presents the information-theoretic bounds for prediction. Concluding remarks are given in Section~5.

\section{Preliminaries}

In this paper, we consider real-valued continuous random variables and vectors, as well as discrete-time stochastic processes they compose. We assume that the support sets of the random variables and vectors are compact if they are bounded. All the random variables, random vectors, and stochastic processes are assumed to be zero-mean, for simplicity and without loss of generality (see Subsection~\ref{maxdeviation} for detailed discussions). We represent random variables and vectors using boldface letters. Given a stochastic process $\left\{ \mathbf{x}_{k}\right\}, \mathbf{x}_{k} \in \mathbb{R}$, we denote the sequence $\mathbf{x}_0,\ldots,\mathbf{x}_{k}$ by the random vector $\mathbf{x}_{0,\ldots,k}=\left[\mathbf{x}_0,\ldots,\mathbf{x}_{k}\right]^T$ for simplicity. The logarithm is defined with base $2$. All functions are assumed to be measurable. 
A stochastic process $\left\{ \mathbf{x}_{k}\right\}, \mathbf{x}_{k} \in \mathbb{R}$ is said to be asymptotically stationary if it is stationary as $k \to \infty$, and herein stationarity means strict stationarity unless otherwise specified \cite{Pap:02}. In addition, a process being asymptotically stationary implies that it is asymptotically mean stationary \cite{gray2011entropy}.

Definitions and properties of the information-theoretic notions that will be used in this paper, including differential entropy $h\left( \mathbf{x} \right)$, conditional differential entropy $
h\left(\mathbf{x} \middle| \mathbf{y}\right)$, entropy rate $h_\infty \left(\mathbf{x}\right)$, and mutual information $I\left(\mathbf{x};\mathbf{y}\right)$, can be found in, e.g., \cite{Cov:06}. In particular, the next lemma \cite{Cov:06} shows that the uniform distribution is the maximum entropy distribution over a finite support.

\begin{lemma} \label{maximum}
	Consider a random variable $\mathbf{x} \in \mathbb{R}$ with (compact) support $\mathrm{supp} \left( \mathbf{x} \right)$. Denote the length of the support as $\left| \mathrm{supp} \left( \mathbf{x} \right) \right|$, and assume that $0 < \left| \mathrm{supp} \left( \mathbf{x} \right) \right| < \infty$.
	Then,  
	\begin{flalign} 
	h \left( \mathbf{x} \right) \leq \log \left| \mathrm{supp} \left( \mathbf{x} \right) \right|, \nonumber
	\end{flalign}
	where equality holds if and only if $\mathbf{x}$ is uniformly distributed over the support (in such a case, we say that $\mathbf{x}$ is uniform for simplicity in the rest of the paper).
\end{lemma}

\section{Bounds for Estimation} \label{estimation}

We first provide a generic bound on the support length of the estimation error's distribution for when estimating a sequence with side information.

\begin{theorem} \label{MIMOFano}
	Consider a stochastic process $\left\{ \mathbf{x}_{k} \right\}, \mathbf{x}_{k} \in \mathbb{R}$ with side information $\left\{ \mathbf{y}_{k} \right\}$. Denote the estimation of $\mathbf{x}_{k}$ by $\widehat{\mathbf{x}}_{k} = f_{k} \left( \mathbf{y}_{0,\ldots,k} \right)$.
	Then,  
	\begin{flalign} \label{MIMOestimation}
	\left| \mathrm{supp} \left( \mathbf{x}_{k} - \widehat{\mathbf{x}}_{k} \right) \right|
	\geq 2^{h \left( \mathbf{x}_k | \mathbf{y}_{0,\ldots,k} \right)},
	\end{flalign}
	where equality holds if and only if $\mathbf{x}_{k} - \widehat{\mathbf{x}}_{k}$ is uniform and $I \left( \mathbf{x}_k - \widehat{\mathbf{x}}_{k}; \mathbf{y}_{0,\ldots,k} \right) = 0$.
\end{theorem}

{\it Proof.}
	It is known from Lemma~\ref{maximum} that 
	\begin{flalign} \label{infinite}
	\left| \mathrm{supp} \left( \mathbf{x}_{k} - \widehat{\mathbf{x}}_{k} \right) \right|
	\geq 2^{ h \left( \mathbf{x}_k - \widehat{\mathbf{x}}_{k} \right)}, 
	\end{flalign} 
	where equality holds if and only if $\mathbf{x}_{k} - \widehat{\mathbf{x}}_{k}$ is uniform. Meanwhile,
	\begin{flalign} 
	&h \left( \mathbf{x}_k - \widehat{\mathbf{x}}_{k} \right)
	= h \left( \mathbf{x}_k - \widehat{\mathbf{x}}_{k} | \mathbf{y}_{0,\ldots,k} \right) + I \left( \mathbf{x}_k - \widehat{\mathbf{x}}_{k}; \mathbf{y}_{0,\ldots,k} \right) \nonumber \\
	&~~~~ = h \left( \mathbf{x}_k - f_{k} \left( \mathbf{y}_{0,\ldots,k} \right) | \mathbf{y}_{0,\ldots,k} \right) + I \left( \mathbf{x}_k - \widehat{\mathbf{x}}_{k}; \mathbf{y}_{0,\ldots,k} \right) \nonumber \\
	&~~~~ = h \left( \mathbf{x}_k | \mathbf{y}_{0,\ldots,k} \right) + I \left( \mathbf{x}_k - \widehat{\mathbf{x}}_{k}; \mathbf{y}_{0,\ldots,k} \right) 
	\geq h \left( \mathbf{x}_k | \mathbf{y}_{0,\ldots,k} \right). \nonumber
	\end{flalign}
	As a result,
	$
	2^{ h \left( \mathbf{x}_k - \widehat{\mathbf{x}}_{k} \right)} 
	\geq 2^{h \left( \mathbf{x}_k | \mathbf{y}_{0,\ldots,k} \right)},
	$
	where equality holds if and only if $I \left( \mathbf{x}_k - \widehat{\mathbf{x}}_{k}; \mathbf{y}_{0,\ldots,k} \right) = 0$.
	Therefore,
	$
	\left| \mathrm{supp} \left( \mathbf{x}_{k} - \widehat{\mathbf{x}}_{k} \right) \right|
	\geq 2^{h \left( \mathbf{x}_k | \mathbf{y}_{0,\ldots,k} \right)}, 
	$ 
	where equality holds if and only if $\mathbf{x}_{k} - \widehat{\mathbf{x}}_{k}$ is uniform and $I \left( \mathbf{x}_k - \widehat{\mathbf{x}}_{k}; \mathbf{y}_{0,\ldots,k} \right) = 0$. This completes the proof.
\hfill$\square$
 
Herein, the lower bound is determined completely by the conditional entropy of the data point $\mathbf{x}_{k}$ to be estimated conditioned on the side information $\mathbf{y}_{0,\ldots,k}$, that is, the amount of ``uncertainty" contained in $\mathbf{x}_{k}$ given $\mathbf{y}_{0,\ldots,k}$. In addition, if $\mathbf{y}_{0,\ldots,k}$ provides more/less information of $\mathbf{x}_{k}$, then the conditional entropy becomes smaller/larger, and thus the bound becomes smaller/larger. In the limit when $\mathbf{x}_{k}$ is independent of $\mathbf{y}_{0,\ldots,k}$, we have $h \left( \mathbf{x}_k | \mathbf{y}_{0,\ldots,k} \right) = h \left( \mathbf{x}_k \right)$, and thus
$
\left| \mathrm{supp} \left( \mathbf{x}_{k} - \widehat{\mathbf{x}}_{k} \right) \right|
\geq 2^{h \left( \mathbf{x}_k \right)}. 
$

It is clear that equality in \eqref{MIMOestimation} holds if and only if the estimation error $\mathbf{x}_k - \widehat{\mathbf{x}}_{k}$ is uniform and contains no information of the side information $\mathbf{y}_{0,\ldots,k}$; it is as if all the ``information" that may be utilized to reduce the estimation error's support length has been extracted.

In \eqref{infinite}, even with a finite $h \left( \mathbf{x}_k - \widehat{\mathbf{x}}_{k} \right)$, $\left| \mathrm{supp} \left( \mathbf{x}_{k} - \widehat{\mathbf{x}}_{k} \right) \right|$ can be infinite if the support set of $\mathbf{x}_{k} - \widehat{\mathbf{x}}_{k} $ is unbounded (e.g., when it is Gaussian); though in such cases \eqref{MIMOestimation} still holds, this means there exists no finite upper bound on $\left| \mathrm{supp} \left( \mathbf{x}_{k} - \widehat{\mathbf{x}}_{k} \right) \right|$.

\subsection{Bounds on the maximum estimation deviations} \label{maxdeviation}

We now examine how the bound on the support length of the estimation error can be related to its maximum deviation $\mathrm{D}_{\max} \left( \mathbf{x}_{k} - \widehat{\mathbf{x}}_{k} \right)$, defined as
\begin{flalign} \label{deviationdef}
\mathrm{D}_{\max} \left( \mathbf{x}_{k} - \widehat{\mathbf{x}}_{k} \right)
\triangleq \max_{ \left( \mathbf{x}_{k} - \widehat{\mathbf{x}}_{k} \right) \in \mathrm{supp} \left( \mathbf{x}_{k} - \widehat{\mathbf{x}}_{k} \right) } \left| \mathbf{x}_{k} - \widehat{\mathbf{x}}_{k} \right|,
\end{flalign}
and denoting the maximum (absolute) deviation of $ \mathbf{x}_{k} - \widehat{\mathbf{x}}_{k} $ from its mean (which is zero since $\mathbf{x}_{k}$ and $ \widehat{\mathbf{x}}_{k}$ are assumed zero-mean by default).
In particular, it can be shown that
\begin{flalign} \label{volume}
\mathrm{D}_{\max} \left( \mathbf{x}_{k} - \widehat{\mathbf{x}}_{k} \right)
\geq \frac{\left| \mathrm{supp} \left( \mathbf{x}_{k} - \widehat{\mathbf{x}}_{k} \right) \right|}{2},
\end{flalign}
where equality holds if and only if $\mathbf{x}_{k} - \widehat{\mathbf{x}}_{k}$ has a symmetric distribution.
Note that herein, with a slight abuse of notation, the $\left| \cdot \right| $ in \eqref{volume} denotes the volume of the support, while that in \eqref{deviationdef} denotes the absolute value. 

In fact, \eqref{volume} holds as long as $\mathbf{x}_{k} - \widehat{\mathbf{x}}_{k}$ is zero-mean, i.e., $\widehat{\mathbf{x}}_{k}$ is an unbiased estimate of $\mathbf{x}_{k}$. Note that if $\mathbf{x}_{k} - \widehat{\mathbf{x}}_{k}$ is not zero-mean, then the maximum deviation is defined as
\begin{flalign} 
&\mathrm{D}_{\max} \left( \mathbf{x}_{k} - \widehat{\mathbf{x}}_{k} \right) \nonumber \\
&~~~~ \triangleq \max_{ \left( \mathbf{x}_{k} - \widehat{\mathbf{x}}_{k} \right) \in \mathrm{supp} \left( \mathbf{x}_{k} - \widehat{\mathbf{x}}_{k} \right) } \left| \mathbf{x}_{k} - \widehat{\mathbf{x}}_{k} - \mathbb{E} \left( \mathbf{x}_{k} - \widehat{\mathbf{x}}_{k} \right) \right|,
\end{flalign}
where $\mathbb{E} \left( \mathbf{x}_{k} - \widehat{\mathbf{x}}_{k} \right)$ denotes the mean of $ \mathbf{x}_{k} - \widehat{\mathbf{x}}_{k} $. In such a case, it still holds that 
\begin{flalign} 
\mathrm{D}_{\max} \left( \mathbf{x}_{k} - \widehat{\mathbf{x}}_{k} \right)
\geq \frac{\left| \mathrm{supp} \left( \mathbf{x}_{k} - \widehat{\mathbf{x}}_{k} \right) \right|}{2},
\end{flalign}
where equality holds if and only if $\mathbf{x}_{k} - \widehat{\mathbf{x}}_{k}$ has a symmetric distribution. In this sense, concerning the bounds derived in this paper, there is no loss of generality when it is assumed that $\mathbf{x}_{k}$ and $\widehat{\mathbf{x}}_{k}$ are zero-mean.

In what follows, we may then derive the corresponding bound on the maximum deviation simply using \eqref{volume}.


\begin{theorem} \label{deviation}
	Consider a stochastic process $\left\{ \mathbf{x}_{k} \right\}, \mathbf{x}_{k} \in \mathbb{R}$ with side information $\left\{ \mathbf{y}_{k} \right\}$. Denote the estimation of $\mathbf{x}_{k}$ by $\widehat{\mathbf{x}}_{k} = f_{k} \left( \mathbf{y}_{0,\ldots,k} \right)$.
	Then,  
	\begin{flalign}
	\mathrm{D}_{\max} \left( \mathbf{x}_{k} - \widehat{\mathbf{x}}_{k} \right)
	\geq 2^{h \left( \mathbf{x}_k | \mathbf{y}_{0,\ldots,k} \right) - 1},
	\end{flalign}
	where equality holds if and only if $\mathbf{x}_{k} - \widehat{\mathbf{x}}_{k}$ is uniform and $I \left( \mathbf{x}_k - \widehat{\mathbf{x}}_{k}; \mathbf{y}_{0,\ldots,k} \right) = 0$.
\end{theorem}

{\it Proof.} Combining Theorem~\ref{MIMOFano} and \eqref{volume}, it follows that
\begin{flalign}
\mathrm{D}_{\max} \left( \mathbf{x}_{k} - \widehat{\mathbf{x}}_{k} \right)
\geq \frac{\left| \mathrm{supp} \left( \mathbf{x}_{k} - \widehat{\mathbf{x}}_{k} \right) \right|}{2}
\geq 2^{h \left( \mathbf{x}_k | \mathbf{y}_{0,\ldots,k} \right) - 1}. \nonumber
\end{flalign}	
Since uniform distributions are symmetric, we have
\begin{flalign}
\mathrm{D}_{\max} \left( \mathbf{x}_{k} - \widehat{\mathbf{x}}_{k} \right)
= 2^{h \left( \mathbf{x}_k | \mathbf{y}_{0,\ldots,k} \right) - 1} \nonumber
\end{flalign}
if and only if $\mathbf{x}_{k} - \widehat{\mathbf{x}}_{k}$ is uniform and $I \left( \mathbf{x}_k - \widehat{\mathbf{x}}_{k}; \mathbf{y}_{0,\ldots,k} \right) = 0$. This completes the proof.
\hfill$\square$


\subsection{Generality of the bounds}

Note that in the estimation (and the subsequent prediction) bounds obtained in this paper, no specific restrictions on the classes of learning algorithms that can be applied have been imposed. Simply put, the bounds are valid for arbitrary learning algorithms in practical use, from classical regression methods to deep learning. On the other hand, the distributions of the sequences to be estimated (or predicted) are, in general, not restricted either; in other words, the sequences could be uniform or non-uniform, white or colored, stationary or non-stationary, and so on. 

The generic performance bounds provide baselines for performance assessment and evaluation of various machine learning and data analysis algorithms. Such baselines function as fundamental benchmarks that separate what is possible and what is impossible, and can thus be applied to indicate how much room is left for performance improvement in learning algorithm design, or to avoid infeasible design specifications in the first place, saving time to be spent on unnecessary parameter tuning work that is destined to be futile.

\section{Bounds for Prediction}

In what follows, we introduce a generic bound on the support length of the prediction error for when predicting a data point based on its previous data points. 

\begin{theorem} \label{MIMOprediction}
	Consider a stochastic process $\left\{ \mathbf{x}_{k} \right\}, \mathbf{x}_{k} \in \mathbb{R}$. Denote the $1$-step ahead prediction (in the rest of the paper, ``$1$-step ahead prediction" will be abbreviated as ``$1$-step prediction" for simplicity) of $\mathbf{x}_{k}$ by $\widehat{\mathbf{x}}_{k} = f_{k} \left( \mathbf{x}_{0,\ldots,k-1} \right)$.
	Then,  
	\begin{flalign} \label{MIMOprediction1}
	\left| \mathrm{supp} \left( \mathbf{x}_{k} - \widehat{\mathbf{x}}_{k} \right) \right|
	\geq 2^{ h \left( \mathbf{x}_k | \mathbf{x}_{0,\ldots,k-1} \right)},
	\end{flalign}
	where equality holds if and only if $\mathbf{x}_{k} - \widehat{\mathbf{x}}_{k}$ is uniform and $I \left( \mathbf{x}_k - \widehat{\mathbf{x}}_{k}; \mathbf{x}_{0,\ldots,k-1} \right) = 0$.
\end{theorem}

{\it Proof.}
Theorem~\ref{MIMOprediction} follows directly from Theorem~\ref{MIMOFano} by letting $\mathbf{y}_{0,\ldots,k} = \mathbf{x}_{0,\ldots,k-1}$ therein.
\hfill$\square$


Herein, the prediction bound depends only on the conditional entropy of the data point $\mathbf{x}_{k}$ to be predicted given the previous data points $\mathbf{x}_{0,\ldots,k-1}$. 

On the other hand, equality in \eqref{MIMOprediction1} holds if and only if the prediction error $\mathbf{x}_k - \widehat{\mathbf{x}}_{k} $ is uniform, and contains no information of the previous data points $\mathbf{x}_{0,\ldots,k-1}$. In addition, we next present an alternative perspective to view the term $I \left( \mathbf{x}_k - \widehat{\mathbf{x}}_{k}; \mathbf{x}_{0,\ldots,k-1} \right)$. 

\begin{proposition} Assuming that $\widehat{\mathbf{x}}_{k} = f_{k} \left( \mathbf{x}_{0,\ldots,k-1} \right)$, it always holds that
	\begin{flalign}
	&I \left( \mathbf{x}_k - \widehat{\mathbf{x}}_{k}; \mathbf{x}_{0,\ldots,k-1} \right) \nonumber \\
	&~~~~ = I \left( \mathbf{x}_{k} - \widehat{\mathbf{x}}_{k} ; \mathbf{x}_{0} - \widehat{\mathbf{x}}_{0}, \ldots, \mathbf{x}_{k-1} - \widehat{\mathbf{x}}_{k-1} \right).
	\end{flalign}
\end{proposition}

{\it Proof.}
    Since $\widehat{\mathbf{x}}_{k-1} = f_{k-1} \left( \mathbf{x}_{0,\ldots,k-2} \right) $, we have (by the data processing inequality \cite{Cov:06})
	\begin{flalign} 
	&I \left( \mathbf{x}_{k} - \widehat{\mathbf{x}}_{k} ; \mathbf{x}_{0,\ldots,k-1} \right) \nonumber \\
	&~~~~ = I \left( \mathbf{x}_{k} - \widehat{\mathbf{x}}_{k} ; \mathbf{x}_{0,\ldots,k-2}, \mathbf{x}_{k-1} - \widehat{\mathbf{x}}_{k-1} \right).\nonumber
	\end{flalign}
	As such, by invoking the data processing inequality repeatedly, it follows that
	\begin{flalign} 
	&I \left( \mathbf{x}_{k} - \widehat{\mathbf{x}}_{k} ; \mathbf{x}_{0,\ldots,k-2}, \mathbf{x}_{k-1} - \widehat{\mathbf{x}}_{k-1} \right) \nonumber \\
	&~~~~ = I \left( \mathbf{x}_{k} - \widehat{\mathbf{x}}_{k} ; \mathbf{x}_{0,\ldots,k-3}, \mathbf{x}_{k-2} - \widehat{\mathbf{x}}_{k-2}, \mathbf{x}_{k-1} - \widehat{\mathbf{x}}_{k-1} \right) 
	\nonumber \\
	&~~~~ = \cdots = I \left( \mathbf{x}_{k} - \widehat{\mathbf{x}}_{k} ; \mathbf{x}_{0} - \widehat{\mathbf{x}}_{0}, \ldots, \mathbf{x}_{k-1} - \widehat{\mathbf{x}}_{k-1} \right). \nonumber
	\end{flalign}
	Eventually, this leads to
	\begin{flalign} 
	&I \left( \mathbf{x}_{k} - \widehat{\mathbf{x}}_{k} ; \mathbf{x}_{0,\ldots,k-1} \right) \nonumber \\
	&~~~~ =I \left( \mathbf{x}_{k} - \widehat{\mathbf{x}}_{k} ; \mathbf{x}_{0} - \widehat{\mathbf{x}}_{0}, \ldots, \mathbf{x}_{k-1} - \widehat{\mathbf{x}}_{k-1} \right), \nonumber
	\end{flalign}
	and completes the proof.
\hfill$\square$

Stated alternatively, the mutual information between the current prediction error and the previous data points is equal to that between the current prediction error and the previous prediction errors.
Accordingly, the condition that 
$
I \left( \mathbf{x}_k - \widehat{\mathbf{x}}_{k}; \mathbf{x}_{0,\ldots,k-1} \right) = 0
$
is equivalent to that 
\begin{flalign}
I \left( \mathbf{x}_{k} - \widehat{\mathbf{x}}_{k} ; \mathbf{x}_{0} - \widehat{\mathbf{x}}_{0}, \ldots, \mathbf{x}_{k-1} - \widehat{\mathbf{x}}_{k-1} \right) = 0,
\end{flalign}
which in turn means that the current prediction error $\mathbf{x}_k - \widehat{\mathbf{x}}_{k} $ contains no information of the previous prediction errors. This is a key link that facilitates the subsequent analysis in the asymptotic case.

\begin{corollary} \label{MIMOasymp}
	Consider a stochastic process $\left\{ \mathbf{x}_{k} \right\}, \mathbf{x}_{k} \in \mathbb{R}$. Denote the $1$-step prediction of $\mathbf{x}_{k}$ by $\widehat{\mathbf{x}}_{k} = f_{k} \left( \mathbf{x}_{0,\ldots,k-1} \right)$.
	Then,  
	\begin{flalign} \label{MIMOasymp1}
	\liminf_{k\to \infty} \left| \mathrm{supp} \left( \mathbf{x}_{k} - \widehat{\mathbf{x}}_{k} \right) \right| 
	\geq \liminf_{k\to \infty} 2^{ h \left( \mathbf{x}_k | \mathbf{x}_{0,\ldots,k-1} \right)},
	\end{flalign}
	where equality holds if $\left\{ \mathbf{x}_{k} - \widehat{\mathbf{x}}_{k} \right\}$ is asymptotically white uniform.
\end{corollary}

{\it Proof.}
	As in the proof of Theorem~\ref{MIMOFano}, it can be shown that
	\begin{flalign} 
	\left| \mathrm{supp} \left( \mathbf{x}_{k} - \widehat{\mathbf{x}}_{k} \right) \right|
	\geq 2^{h \left( \mathbf{x}_k | \mathbf{x}_{0,\ldots,k-1} \right)}, \nonumber
	\end{flalign} 
	where equality holds if and only if $\mathbf{x}_{k} - \widehat{\mathbf{x}}_{k}$ is uniform and $I \left( \mathbf{x}_k - \widehat{\mathbf{x}}_{k}; \mathbf{x}_{0,\ldots,k-1} \right) = 0$. This, by taking $\liminf_{k\to \infty}$ on both sides, then leads to
	\begin{flalign} 
	\liminf_{k\to \infty} \left| \mathrm{supp} \left( \mathbf{x}_{k} - \widehat{\mathbf{x}}_{k} \right) \right|
	\geq \liminf_{k\to \infty} 2^{h \left( \mathbf{x}_k | \mathbf{x}_{0,\ldots,k-1} \right)}. \nonumber
	\end{flalign}
	Herein, equality holds if $\mathbf{x}_{k} - \widehat{\mathbf{x}}_{k}$ is uniform and 
	\begin{flalign}
	&I \left( \mathbf{x}_k - \widehat{\mathbf{x}}_{k}; \mathbf{x}_{0,\ldots,k-1} \right) \nonumber \\
	&~~~~ = I \left( \mathbf{x}_{k} - \widehat{\mathbf{x}}_{k} ; \mathbf{x}_{0} - \widehat{\mathbf{x}}_{0}, \ldots, \mathbf{x}_{k-1} - \widehat{\mathbf{x}}_{k-1} \right)= 0, \nonumber
	\end{flalign}
	as $k\to \infty$. Since $\mathbf{x}_{k} - \widehat{\mathbf{x}}_{k}$ being uniform as $k\to \infty$ means that $\mathbf{x}_{k} - \widehat{\mathbf{x}}_{k}$ is asymptotically uniform, and that 
	\begin{flalign}
	I \left( \mathbf{x}_{k} - \widehat{\mathbf{x}}_{k} ; \mathbf{x}_{0} - \widehat{\mathbf{x}}_{0}, \ldots, \mathbf{x}_{k-1} - \widehat{\mathbf{x}}_{k-1} \right)= 0 \nonumber
	\end{flalign}
	as $k\to \infty$ is equivalent to that $\mathbf{x}_{k} - \widehat{\mathbf{x}}_{k}$ is asymptotically white, equality in \eqref{MIMOasymp1} holds if $\left\{ \mathbf{x}_{k} - \widehat{\mathbf{x}}_{k} \right\}$ is asymptotically white uniform.
\hfill$\square$


When the sequence to be predicted is asymptotically stationary, we arrive at the following result.

\begin{corollary} \label{uniform}
	Consider an asymptotically stationary stochastic process $\left\{ \mathbf{x}_{k} \right\}, \mathbf{x}_{k} \in \mathbb{R}$. Denote the $1$-step prediction of $\mathbf{x}_{k}$ by $\widehat{\mathbf{x}}_{k} = f_{k} \left( \mathbf{x}_{0,\ldots,k-1} \right)$.
	Then,
	\begin{flalign} 
	\liminf_{k\to \infty} \left| \mathrm{supp} \left( \mathbf{x}_{k} - \widehat{\mathbf{x}}_{k} \right) \right| 
	\geq 2^{ h_{\infty} \left( \mathbf{x} \right)},
	\end{flalign}
	where $h_{\infty} \left( \mathbf{x} \right)$ denotes the entropy rate \cite{Cov:06} of $\left\{ \mathbf{x}_{k} \right\}$. Herein, equality holds if $\left\{ \mathbf{x}_{k} - \widehat{\mathbf{x}}_{k} \right\}$ is asymptotically white uniform.
\end{corollary}

{\it Proof.}
Corollary~\ref{uniform} follows directly from Corollary~\ref{MIMOasymp} by noting that for an asymptotically stationary process $\left\{ \mathbf{x}_{k} \right\}$, we have \cite{Cov:06} 
\begin{flalign} 
\liminf_{k\to \infty} h \left( \mathbf{x}_k | \mathbf{x}_{0,\ldots,k-1} \right) 
= \lim_{k\to \infty} h \left( \mathbf{x}_k | \mathbf{x}_{0,\ldots,k-1} \right)
= h_{\infty} \left( \mathbf{x} \right). \nonumber
\end{flalign}
This completes the proof.
\hfill$\square$

As a matter of fact, all the previous bounds
are derived for $1$-step prediction. For the more general case
of $m$-step prediction (where $m$ is any positive integer), the
following bound can be obtained.

\begin{theorem} \label{mstep}
	Consider a stochastic process $\left\{ \mathbf{x}_{k} \right\}, \mathbf{x}_{k} \in \mathbb{R}$. Denote the $m$-step prediction of $\mathbf{x}_{k}$ by $\widehat{\mathbf{x}}_{k} = f_{k} \left( \mathbf{x}_{0,\ldots,k-m} \right)$.
	Then,  
	\begin{flalign} 
	\left| \mathrm{supp} \left( \mathbf{x}_{k} - \widehat{\mathbf{x}}_{k} \right) \right|
	\geq 2^{ h \left( \mathbf{x}_k | \mathbf{x}_{0,\ldots,k-m} \right)},
	\end{flalign}
	where equality holds if and only if $\mathbf{x}_{k} - \widehat{\mathbf{x}}_{k}$ is uniform and $I \left( \mathbf{x}_k - \widehat{\mathbf{x}}_{k}; \mathbf{x}_{0,\ldots,k-m} \right) = 0$.
\end{theorem}

We now examine further the term $h \left( \mathbf{x}_k | \mathbf{x}_{0,\ldots,k-m} \right)$.

\begin{proposition}
	It always holds that
	\begin{flalign} 
	h \left( \mathbf{x}_k | \mathbf{x}_{0,\ldots,k-1} \right)
	&\leq \cdots \leq h \left( \mathbf{x}_k | \mathbf{x}_{0,\ldots,k-m+1} \right) \nonumber \\
	&\leq h \left( \mathbf{x}_k | \mathbf{x}_{0,\ldots,k-m} \right) \leq h \left( \mathbf{x}_k \right).
	\end{flalign}
\end{proposition}

{\it Proof.}
To compare with the $l$-step (where $1 \leq l < m$) prediction bound, note that
\begin{flalign} 
h \left( \mathbf{x}_k | \mathbf{x}_{0,\ldots,k-m} \right) 
&= h \left( \mathbf{x}_k | \mathbf{x}_{0,\ldots,k-l} \right) \nonumber \\
&~~~~ + I \left( \mathbf{x}_k ; \mathbf{x}_{k-m+1,\ldots,k-l} | \mathbf{x}_{0,\ldots,k-m} \right). \nonumber
\end{flalign}
As such,
\begin{flalign} 
h \left( \mathbf{x}_k | \mathbf{x}_{0,\ldots,k-m} \right) 
&\geq h \left( \mathbf{x}_k | \mathbf{x}_{0,\ldots,k-m+1} \right) \nonumber \\
&\geq \cdots \geq h \left( \mathbf{x}_k | \mathbf{x}_{0,\ldots,k-1} \right). \nonumber
\end{flalign}
On the other hand,
\begin{flalign} 
h \left( \mathbf{x}_k | \mathbf{x}_{0,\ldots,k-m} \right) 
= h \left( \mathbf{x}_k \right) - I \left( \mathbf{x}_k ; \mathbf{x}_{0,\ldots,k-m} \right), \nonumber
\end{flalign}
and hence
$ 
h \left( \mathbf{x}_k | \mathbf{x}_{0,\ldots,k-m} \right) 
\leq h \left( \mathbf{x}_k \right).
$
\hfill$\square$

That is to say, the prediction bound will not decrease as the prediction step increases. Additionally, in the worst case, the bound is given by $2^{h \left( \mathbf{x}_k \right)}$.

When $k \to \infty$, the next corollary follows for $m$-step prediction.

\begin{corollary} \label{asymmstep}
	Consider a stochastic process $\left\{ \mathbf{x}_{k} \right\}, \mathbf{x}_{k} \in \mathbb{R}$. Denote the $m$-step prediction of $\mathbf{x}_{k}$ by $\widehat{\mathbf{x}}_{k} = f_{k} \left( \mathbf{x}_{0,\ldots,k-m} \right)$.
	Then,  
	\begin{flalign} \label{MIMOasymp1m}
	\liminf_{k\to \infty} \left| \mathrm{supp} \left( \mathbf{x}_{k} - \widehat{\mathbf{x}}_{k} \right) \right| 
	\geq \liminf_{k\to \infty} 2^{ h \left( \mathbf{x}_k | \mathbf{x}_{0,\ldots,k-m} \right)},
	\end{flalign}
	where equality holds if $\left\{ \mathbf{x}_{k} - \widehat{\mathbf{x}}_{k} \right\}$ is asymptotically uniform and colored up to the order of $m-1$.
\end{corollary}

Herein, a stochastic process is said to be colored up to the order of $m-1$ if $\mathbf{x}_{k}$ is independent of $\mathbf{x}_{k-m,k-m-1, \ldots}$. Clearly, when $m=1$, being colored up to order $m-1$ is equivalent to being white, which reduces to the case of $1$-step prediction.

\subsection{Bounds on the maximum prediction deviations}

Since 
\begin{flalign}
\mathrm{D}_{\max} \left( \mathbf{x}_{k} - \widehat{\mathbf{x}}_{k} \right)
&= \max_{ \left( \mathbf{x}_{k} - \widehat{\mathbf{x}}_{k} \right) \in \mathrm{supp} \left( \mathbf{x}_{k} - \widehat{\mathbf{x}}_{k} \right) } \left| \mathbf{x}_{k} - \widehat{\mathbf{x}}_{k} \right| \nonumber \\
&\geq \frac{\left| \mathrm{supp} \left( \mathbf{x}_{k} - \widehat{\mathbf{x}}_{k} \right) \right|}{2},
\end{flalign}
corresponding bounds for $\mathrm{D}_{\max} \left( \mathbf{x}_{k} - \widehat{\mathbf{x}}_{k} \right)$ may then be obtained based on the prediction bounds derived previously in this section (cf. analysis and deviations in Subsection~\ref{maxdeviation}). For instance, the counterpart to Theorem~\ref{MIMOprediction} is presented as follows.

\begin{theorem} \label{md1}
	Consider a stochastic process $\left\{ \mathbf{x}_{k} \right\}, \mathbf{x}_{k} \in \mathbb{R}$. Denote the $1$-step prediction of $\mathbf{x}_{k}$ by $\widehat{\mathbf{x}}_{k} = f_{k} \left( \mathbf{x}_{0,\ldots,k-1} \right)$.
	Then,  
	\begin{flalign}
	\mathrm{D}_{\max} \left( \mathbf{x}_{k} - \widehat{\mathbf{x}}_{k} \right)
	\geq 2^{h \left( \mathbf{x}_k | \mathbf{x}_{0,\ldots,k-1} \right) - 1}.
	\end{flalign}
	Herein, equality holds if and only if $\mathbf{x}_{k} - \widehat{\mathbf{x}}_{k}$ is uniform and $I \left( \mathbf{x}_k - \widehat{\mathbf{x}}_{k}; \mathbf{x}_{0,\ldots,k-1} \right) = 0$.
\end{theorem}

In the asymptotic case, the following corollary holds in parallel with Corollary~\ref{MIMOasymp}.

\begin{corollary} \label{md2}
	Consider a stochastic process $\left\{ \mathbf{x}_{k} \right\}, \mathbf{x}_{k} \in \mathbb{R}$. Denote the $1$-step prediction of $\mathbf{x}_{k}$ by $\widehat{\mathbf{x}}_{k} = f_{k} \left( \mathbf{x}_{0,\ldots,k-1} \right)$.
	Then,  
	\begin{flalign} 
	\liminf_{k\to \infty} \mathrm{D}_{\max} \left( \mathbf{x}_{k} - \widehat{\mathbf{x}}_{k} \right)
	\geq \liminf_{k\to \infty} 2^{ h \left( \mathbf{x}_k | \mathbf{x}_{0,\ldots,k-1} \right) - 1},
	\end{flalign}
	where equality holds if $\left\{ \mathbf{x}_{k} - \widehat{\mathbf{x}}_{k} \right\}$ is asymptotically white uniform.
\end{corollary}

Similarly, corresponding results for Corollary~\ref{uniform}, Theorem~\ref{mstep}, as well as Corollary~\ref{asymmstep} can be obtained.

\subsection{Prediction of asymptotically stationary sequences}

Indeed, formulae that are more specific than that of Corollary~\ref{uniform} could be derived when it comes to predicting asymptotically stationary sequences.

\begin{corollary} \label{power1}
	Consider an asymptotically stationary stochastic process $\left\{ \mathbf{x}_{k} \right\}, \mathbf{x}_{k} \in \mathbb{R}$ with asymptotic power spectrum $ S_{\mathbf{x}} \left( \omega \right)$, which is defined as \cite{Pap:02}
	\begin{flalign}
	&S_{\mathbf{x}}\left( \omega\right)
	=\sum_{k=-\infty}^{\infty} R_{\mathbf{x}}\left( k\right) \mathrm{e}^{-\mathrm{j}\omega k}, \nonumber 
	\end{flalign}
	and 
	$R_{\mathbf{x}}\left( k\right) =\lim_{i\to \infty} \mathrm{E}\left[ \mathbf{x}_i \mathbf{x}_{i+k} \right]$ denotes the asymptotic correlation matrix.
	Denote the $1$-step prediction of $\mathbf{x}_{k}$ by $\widehat{\mathbf{x}}_{k} = f_{k} \left( \mathbf{x}_{0,\ldots,k-1} \right)$.
	Then, 	
	\begin{flalign} \label{spectrum}
	&\liminf_{k\to \infty} \left| \mathrm{supp} \left( \mathbf{x}_{k} - \widehat{\mathbf{x}}_{k} \right) \right| \nonumber \\
	&~~~~ \geq \left[ 2^{- J_{\infty} \left( \mathbf{x} \right)} \right] 2^{\frac{1}{2 \mathrm{\pi}}\int_{-\mathrm{\pi}}^{\mathrm{\pi}}{\log \sqrt{2 \pi \mathrm{e} S_{ \mathbf{x} } \left( \omega \right) } \mathrm{d}\omega }},
	\end{flalign}
	where $J_{\infty} \left( \mathbf{x} \right)$ denotes the negentropy rate \cite{fang2017towards} of $\left\{ \mathbf{x}_{k} \right\}$, $J_{\infty} \left( \mathbf{x} \right) \geq 0$, and $J_{\infty} \left( \mathbf{x} \right) = 0$ if and only if $\left\{ \mathbf{x}_{k} \right\}$ is Gaussian.
    Herein, equality holds if $\left\{ \mathbf{x}_{k} - \widehat{\mathbf{x}}_{k} \right\}$ is asymptotically white uniform.
\end{corollary}

{\it Proof.}
It is known from \cite{fang2017towards} that for an asymptotically stationary stochastic process $\left\{ \mathbf{x}_{k} \right\}$ with asymptotic power spectrum $ S_{\mathbf{x}} \left( \omega \right)$,
\begin{flalign} 
h_{\infty} \left( \mathbf{x} \right) 
= \frac{1}{2 \mathrm{\pi}} \int_{-\mathrm{\pi}}^{\mathrm{\pi}} \log \sqrt{2 \pi \mathrm{e} S_{ \mathbf{x} } \left( \omega \right) } \mathrm{d}\omega - J_{\infty} \left( \mathbf{x} \right).  \nonumber
\end{flalign}
Consequently,
\begin{flalign}
2^{h_{\infty} \left( \mathbf{x} \right)} 
= \left[ 2^{- J_{\infty} \left( \mathbf{x} \right)} \right] 2^{\frac{1}{2 \mathrm{\pi}}\int_{-\mathrm{\pi}}^{\mathrm{\pi}}{\log \sqrt{2 \pi \mathrm{e} S_{ \mathbf{x} } \left( \omega \right) } \mathrm{d}\omega }}. \nonumber
\end{flalign}
This completes the proof.
\hfill$\square$

Herein, negentropy rate is a measure of non-Gaussianity for asymptotically stationary sequences, which grows larger as the sequence to be predicted becomes less Gaussian; see \cite{fang2017towards} for more details of its properties. Accordingly, the bounds in \eqref{spectrum} will decrease as $\left\{ \mathbf{x}_{k} \right\}$ becomes less Gaussian, and vice versa. In the limit when  $\left\{ \mathbf{x}_{k} \right\}$ is Gaussian, \eqref{spectrum} reduces to
\begin{flalign} 
\liminf_{k\to \infty} \left| \mathrm{supp} \left( \mathbf{x}_{k} - \widehat{\mathbf{x}}_{k} \right) \right| \geq 2^{\frac{1}{2 \mathrm{\pi}}\int_{-\mathrm{\pi}}^{\mathrm{\pi}}{\log \sqrt{2 \pi \mathrm{e} S_{ \mathbf{x} } \left( \omega \right) } \mathrm{d}\omega }}.
\end{flalign}

Meanwhile, the following bound on the maximum prediction deviation holds.

\begin{corollary} \label{power2}
	Consider an asymptotically stationary stochastic process $\left\{ \mathbf{x}_{k} \right\}, \mathbf{x}_{k} \in \mathbb{R}$ with asymptotic power spectrum $ S_{\mathbf{x}} \left( \omega \right)$.
	Denote the $1$-step prediction of $\mathbf{x}_{k}$ by $\widehat{\mathbf{x}}_{k} = f_{k} \left( \mathbf{x}_{0,\ldots,k-1} \right)$.
	Then, 	
	\begin{flalign} \label{spectrum2}
	&\liminf_{k\to \infty} \mathrm{D}_{\max} \left( \mathbf{x}_{k} - \widehat{\mathbf{x}}_{k} \right) \nonumber \\
	&~~~~ \geq \left[ 2^{- J_{\infty} \left( \mathbf{x} \right)} \right] 2^{\frac{1}{2 \mathrm{\pi}}\int_{-\mathrm{\pi}}^{\mathrm{\pi}}{\log \sqrt{\frac{\pi \mathrm{e}}{2} S_{ \mathbf{x} } \left( \omega \right) } \mathrm{d}\omega }},
	\end{flalign}
	where equality holds if $\left\{ \mathbf{x}_{k} - \widehat{\mathbf{x}}_{k} \right\}$ is asymptotically white uniform.
\end{corollary}

Similarly, when $\left\{ \mathbf{x}_{k} \right\}$ is Gaussian, \eqref{spectrum2} becomes
\begin{flalign} 
\liminf_{k\to \infty} \mathrm{D}_{\max} \left( \mathbf{x}_{k} - \widehat{\mathbf{x}}_{k} \right) \geq 2^{\frac{1}{2 \mathrm{\pi}}\int_{-\mathrm{\pi}}^{\mathrm{\pi}}{\log \sqrt{\frac{\pi \mathrm{e}}{2} S_{ \mathbf{x} } \left( \omega \right) } \mathrm{d}\omega }}.
\end{flalign}

\subsection{``Uniformizing-whitening principle"}

Consider again Corollary~\ref{uniform}. As a matter of fact, if $\left\{ \mathbf{x}_{k} - \overline{\mathbf{x}}_{k} \right\}$ is asymptotically white uniform, then, since $\left\{ \mathbf{x}_{k} \right\}$ is asymptotically stationary, it holds that
\begin{flalign} \label{equality}
\lim_{k\to \infty} \left| \mathrm{supp} \left( \mathbf{x}_{k} - \widehat{\mathbf{x}}_{k} \right) \right| 
= 2^{ h_{\infty} \left( \mathbf{x} \right)}.
\end{flalign}
In addition, we can show that \eqref{equality} holds if and only if $\left\{ \mathbf{x}_{k} - \overline{\mathbf{x}}_{k} \right\}$ is asymptotically white uniform; in other words, the necessary and sufficient condition for achieving the prediction bounds asymptotically is that the prediction error is asymptotically white uniform.  
In other words, the optimal (in the sense of minimizing the support length of the prediction error, or equivalently, minimizing the maximum deviation in the prediction error) learning algorithm is prediction error uniformizing-whitening, which may feature a unifying principle that is applicable to generic sequential prediction problems. 

As mentioned earlier, the generic bounds themselves provide baselines for performance assessment and evaluation of arbitrary sequential prediction algorithms, by comparing the real performance and that given by the theoretical bound. This can be reinforced by observing whether the prediction error is a white uniform process. On the other hand, for future research, we aim to develop sequential learning frameworks that are oriented to the aforementioned fundamental limits in sequential prediction. This is enabled by turning the uniformizing-whitening principle into optimality conditions or even objective functions for the optimization problems formulated accordingly.

\section{Conclusion}
	In this paper, we have derived generic, information-theoretic bounds on the maximum deviations in prediction errors in sequential prediction. The fundamental bounds are applicable to any learning algorithms while the sequences to be predicted can have arbitrary distributions. For future research, we aim to investigate further implications of the bounds as well as the corresponding necessary and sufficient conditions to achieve them.
	
\section{Acknowledgment}
	The authors would like to thank the anonymous reviewers for the many insightful comments and suggestions.
\bibliographystyle{IEEEbib}
\bibliography{references}

\end{document}